\documentclass[journal,twoside,web]{ieeecolor}
\usepackage{generic}

\usepackage{amsmath,amsfonts}
\usepackage{amssymb}
\usepackage{graphicx}
\usepackage{color}
\usepackage{array}
\usepackage{soul}
\usepackage{subfigure}
\usepackage{bm} 
\usepackage{overpic}
\usepackage[switch]{lineno}
\usepackage{multirow,multicol}
\usepackage{booktabs}
\usepackage{algorithm,algorithmic}
\usepackage{hyperref}
\usepackage[backend=biber, style=ieee, maxcitenames=2, mincitenames=1]{biblatex}
\newcommand{\citet}[1]{\Citeauthor{#1} \cite{#1}}
\usepackage{textcomp}
\def\BibTeX{{\rm B\kern-.05em{\sc i\kern-.025em b}\kern-.08em
    T\kern-.1667em\lower.7ex\hbox{E}\kern-.125emX}}
\begin{document}
\title{EvoFA: Evolvable Fast Adaptation for EEG Emotion Recognition}
\author{Ming Jin, Danni Zhang, Gangming Zhao, Changde Du, and Jinpeng Li, \IEEEmembership{Member, IEEE}
\thanks{Manuscript submitted 31 May 2024. This work was supported in part by National Natural Science Foundation of China under Grant 62106248, in part by the Ningbo Clinical Research Center for Medical Imaging under Grant 2021L003, and in part by the Guangdong Provincial Key Laboratory of Human Digital Twin under Grant 2022B1212010004. \emph{(Corresponding author: Jinpeng Li)}}
\thanks{Ming Jin, Danni Zhang, and Jinpeng Li are with Institute of Life and Health Industry, University of Chinese Academy of Sciences, Ningbo 315010, China (e-mail: jinpeng.li@ieee.org).}
\thanks{Gangming Zhao is with Department of Computer Science, University of Hong Kong, Hong Kong 999077, China.}
\thanks{Changde Du is with the Research Center for Brain-Inspired Intelligence, Institute of Automation, Chinese Academy of Sciences, Beijing 100045, China.}
}

\maketitle

\begin{abstract}

Electroencephalography (EEG)-based emotion recognition has gained significant traction due to its accuracy and objectivity. However, the non-stationary nature of EEG signals leads to distribution drift over time, causing severe performance degradation when the model is reused. While numerous domain adaptation (DA) approaches have been proposed in recent years to address this issue, their reliance on large amounts of target data for calibration restricts them to offline scenarios, rendering them unsuitable for real-time applications. To address this challenge, this paper proposes Evolvable Fast Adaptation (EvoFA), an online adaptive framework tailored for EEG data. EvoFA organically integrates the rapid adaptation of Few-Shot Learning (FSL) and the distribution matching of Domain Adaptation (DA) through a two-stage generalization process. During the training phase, a robust base meta-learning model is constructed for strong generalization. In the testing phase, a designed evolvable meta-adaptation module iteratively aligns the marginal distribution of target (testing) data with the evolving source (training) data within a model-agnostic meta-learning framework, enabling the model to learn the evolving trends of testing data relative to training data and improving online testing performance. Experimental results demonstrate that EvoFA achieves significant improvements compared to the basic FSL method and previous online methods. The introduction of EvoFA paves the way for broader adoption of EEG-based emotion recognition in real-world applications. Our code will be released upon publication.
\end{abstract}


\section{Introduction}
\label{sec:introduction}
\IEEEPARstart{E}{motion} recognition holds immense potential for various applications, including education and psychological diagnosis and treatment \cite{wang2020emotion, banskota2023novel}. EEG-based recognition has attracted research interest in recent years due to its low cost, high accuracy, and the difficulty for subjects to conceal their emotions \cite{jimenez2023cross, she2023multisource}. However, despite significant advancements in improving the efficiency and accuracy of recognition, the decline in model performance when reused, especially in online settings, remains a prominent issue limiting the practical application of EEG-based emotion recognition \cite{li2023effective}.

A major obstacle to the reusability of EEG-based emotion recognition is the problem of "domain drift." This issue manifests as changes in the statistical properties of EEG signals over time, significantly impacting the accuracy and reliability of emotion recognition models. Factors contributing to domain drift include changes in the physiological state of subjects (such as fatigue or hunger), environmental interference (such as noise or light), and errors in device usage \cite{zhang2023cross, gong2024ciabl}.

\begin{figure}
  \centering
  \includegraphics[width=0.48\textwidth]{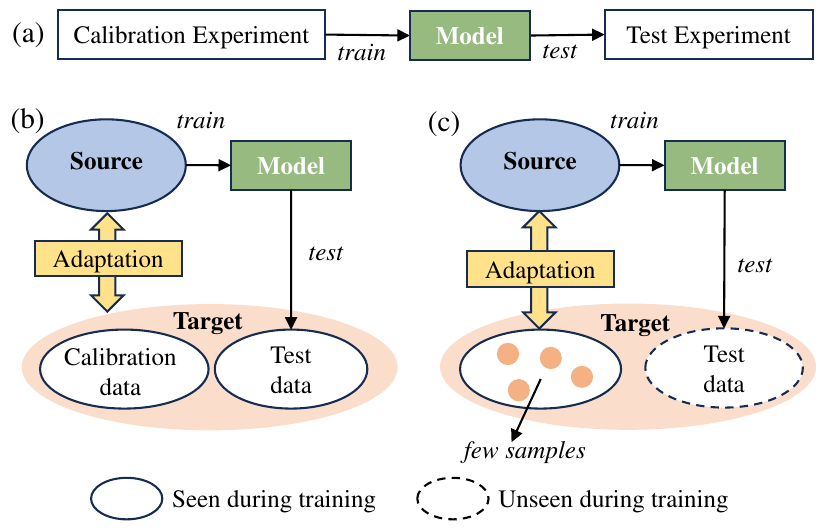}
  \caption{Approaches for adapting EEG emotion recognition models. (a) Calibration-based (supervised) methods employ a significant quantity of labeled calibration data in calibration experiments to retrain or fine-tune the model. (b) Offline adaptation methods utilize all target data, encompassing both calibration and test data, to conduct transductive adaptation. (c) Online adaptation approaches prioritize rapid and efficient model adjustments using a smaller amount of calibration data for inductive adaptation, testing on unseen data.}
  \label{fig1}
\end{figure}

Domain drift poses a significant challenge to model performance . Existing methods to address this issue can be broadly categorized into two approaches.
\begin{itemize}
    \item Supervised methods: As depicted in Figure \ref{fig1}(a), this approach relies on a large amount of labeled data for direct model training or fine-tuning\cite{song2018eeg, jin2023graph}. While it grants generalization capability initially, the model's performance progressively degrades as domain drift in new data becomes persistent. Maintaining high performance necessitates frequent retraining with substantial amounts of newly labeled data, hindering real-world practicality.
    \item Offline adaptation methods: Illustrated in Figure \ref{fig1}(b), this approach utilizes transfer learning to bridge the distributional gap between source and target domains\cite{li2022dynamic, jimenez2023cross}. By leveraging a pre-trained model and fine-tuning it with target data, the models gain discriminative ability on new data. However, existing transfer learning methods still require a significant amount of unlabeled target data for distribution matching, which can be a limitation in resource-constrained scenarios.
\end{itemize}
Both approaches share a common dependency on substantial calibration data for model adjustments. This reliance on large datasets hinders their applicability in real-world situations where data acquisition might be expensive or limited.

Humans exhibit remarkable ability to recognize new categories with only a few examples. Inspired by this capability, FSL has emerged as a promising approach to significantly reduce the calibration data required for models on new data. Applying FSL to EEG-based emotion recognition holds the potential to address the challenge of time-consuming calibration \cite{an2023dual,ning2021cross,zhu2023decoding}. However, the FSL framework primarily focuses on rapid adaptation to entirely new categories, where the training and test sets have non-overlapping in label space. Applying the basic FSL model rigidly to EEG emotion recognition overlooks two critical issues: for EEG data, there may be only limited modality shift between source and target data, and the label space for training and test data are completely identical.

Considering that DA is mainly designed to address data drift from the training set to the test set within the same label space, and FSL enables rapid calibration in an online learning mode, an ideal approach would be to organically combine the online rapid recognition capability of FSL with the distribution matching ability of DA to address data drift.

To address the issue of data drift faced by models during online use, this paper proposes EvoFA, a novel online adaptation framework for EEG-based emotion recognition. As illustrated in Figure \ref{fig1} (c), EvoFA requires only a small amount of calibration data to quickly calibrate the model for effective testing. EvoFA combines the advantages of FSL rapid generalization and DA distribution matching through a two-step generalization process. During the training phase, FSL is utilized to initialize a general emotion recognition model with strong generalization capabilities. In the meta-testing phase, domain shifts induced by data drift are simulated, and an evolvable meta-adaptation module is introduced to align the target domain with the gradually changing source domain, thereby mitigating the impact of data drift on model performance. Experimental results demonstrate that EvoFA achieves significant improvements compared to the basic FSL method and previous online methods.

Our work makes the following significant contributions:
\begin{itemize}
\item We propose EvoFA, a novel framework tailored for online adaptation of EEG data, which significantly reduces calibration data and lowers the barrier for deploying EEG-based emotion recognition in real-world scenarios.
\item Within this framework, we introduce a flexible and lightweight rapid test-time transfer method that mitigates the impact of domain drift caused by continuous changes in EEG data modalities. This method requires no retraining and is compatible with various FSL models.
\item Experimental results on two datasets demonstrate that EvoFA not only achieves superior performance in online recognition tasks but also significantly reduces calibration data.
\end{itemize}

The subsequent sections are organized as follows: Section~\ref{related work} reviews the progress in EEG-based emotion recognition, DA and meta-learning. Section~\ref{methods} primarily introduces our proposed EvoFA framework, which cleverly combines the advantages of FSL rapid generalization and DA distribution matching, making it suitable for EEG emotion recognition tasks. Section~\ref{exp} demonstrates the improvements achieved by EvoFA on intra-subject and inter-subject tasks. Section~\ref{conclusion} summarizes this work and provides future perspectives.

\section{Related Work}\label{related work}

\subsection{EEG Emotion Recognition}

EEG-based emotion recognition can be broadly categorized into three types based on the level of access to test/target data.

\textit{Supervised Learning}: In this approach, test data is entirely unknown. The primary reliance is on supervised learning, where models are trained on training data to develop strong generalization capabilities that can fit well on test data. Since deeper networks typically have stronger fitting abilities, constructing deep models to learn robust representations is a mainstream method \cite{jafari2023emotion, jin2023graph}. In recent years, Graph Convolutional Networks (GCNs) have also been proven effective for handling EEG data due to their ability to capture critical information from brain topologies, allowing better discrimination of emotional patterns at both the feature and brain topology levels \cite{song2018eeg, du2022multi, zhong2020eeg, zeng2022siam}. However, the performance of these trained models often significantly declines when applied to completely unknown test data.

\textit{Offline Learning}: This approach allows access to part or all of the unlabeled test data. To address the inevitable issue of modal drift, many studies have turned to offline learning modes based on transfer learning \cite{li2022dynamic, she2023multisource, jimenez2023cross}. She \emph{et al.} \cite{she2023multisource} proposed a new emotion recognition method based on a multisource associate domain adaptation (DA) network, considering both domain invariant and domain-specific features. SSDA was proposed to align the joint distributions of subjects, assuming that fine-grained structures must be aligned to perform a greater knowledge transfer \cite{jimenez2023cross}. However, since offline learning requires access to a large amount of test data, it often loses its applicability in most real-world scenarios.

\textit{Online Learning}: In this scenario, only a minimal amount of test set calibration data is accessible. To better meet practical application needs, some studies have attempted to achieve rapid calibration and online recognition of EEG data. Pan \emph{et al.} \cite{pan2023multimodal} proposed a novel Online Multimodal Hypergraph Learning (OMHGL) method, which integrates multimodal information based on time-series physiological signals for emotion recognition. Blanco \emph{et al.} \cite{blanco2024real} developed a machine learning model using principal component analysis, power spectral density, random forest, and Extra-Trees that can recognize emotions in real time, providing estimates of valence, arousal, and dominance (VAD) every five seconds. However, to maintain compatibility with real-time requirements, these methods largely rely on traditional machine learning techniques for training.

\subsection{Unsupervised Domain Adaptation}

Unsupervised domain adaptation (UDA) has emerged as a solution to address the insufficient generalization of models across different data distributions. UDA primarily tackles the challenges of costly data collection and annotation by pre-training on existing source data and fine-tuning on unlabeled target data to enhance model performance in the target domain.

UDA methods can be categorized into three main approaches. One approach effectively aligns the distributions of source and target domain data by minimizing the discrepancy in the embedding space, thereby improving model performance on target data, where the domain discrepancy is measured by Maximum Mean Discrepancy (MMD) \cite{long2015learning} and Joint MMD \cite{ long2017deep}. Another common UDA method introduces domain discriminators to train the model to make it difficult for the discriminators to distinguish between source and target domain features, achieving feature alignment \cite{xia2021adaptive, ganin2016domain}. Self-training-based methods generate pseudo-labels for target domain data and utilize these labels to supervise further model training iteratively, progressively enhancing model performance on target data \cite{liu2021cycle, zou2018unsupervised}.

\begin{figure*}[ht]
  \centering
  \includegraphics[width=1.8\columnwidth]{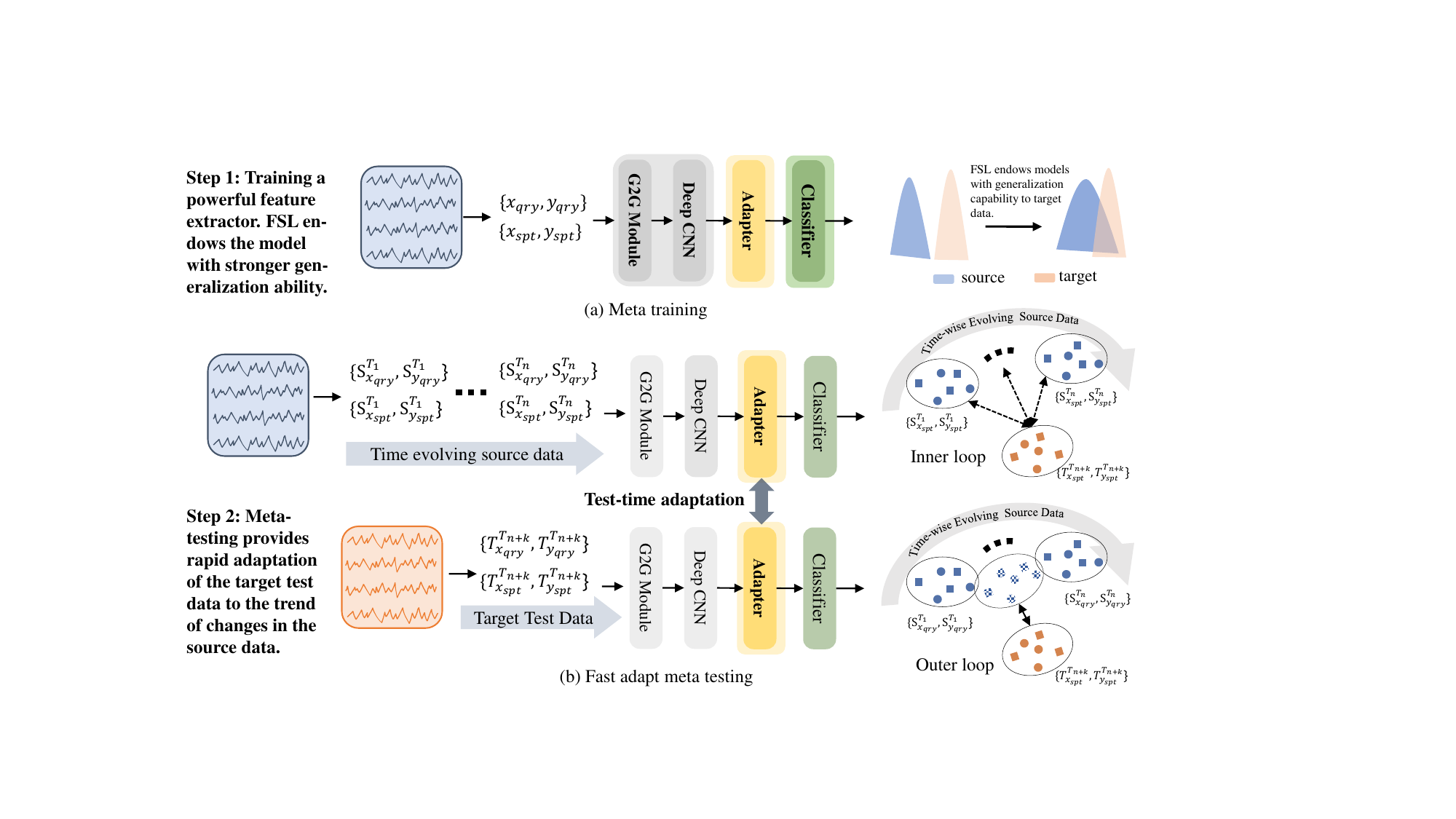}
  \caption{A schematic representation of the two-step generalization in EvoFA.  (a) First generalization (meta-training): initializing an emotion recognition foundation model with strong generalization capabilities based on FSL principles and updating all its parameters to learn these capabilities. (b) Second generalization (fast adapt meta-testing): sample data chronologically from the training dataset to simulate the domain drift of subject data over time. In the inner loop, sequentially reduced the distance between the test data and the support set of the training data. In the outer loop, the model selected an appropriate direction for updating the adapter's parameters. By narrowing the gap with the evolving source, the impact of domain drift on test results can be mitigated.}
  \label{fig2}
\end{figure*}

\subsection{Meta Learning}

Meta-learning, also known as 'learning to learn', stands in contrast to traditional artificial intelligence methods that solve tasks from scratch using fixed learning algorithms. Instead, meta-learning aims to improve the learning algorithm itself based on the experience gained from multiple learning episodes. This approach has gained significant interest in recent years, particularly for tasks involving FSL, which aligns well with the challenges of limited calibration data in EEG-based emotion recognition. The two primary approaches are metric-based and optimization-based approaches.

Metric-based methods focus on learning a meaningful distance or similarity measure, allowing the model to embed data into a space where similar patterns are close together. Common distance metrics include cosine similarity \cite{vinyals2016matching}, Euclidean distance \cite{snell2017prototypical}, and CNN-based relational modules \cite{sung2018learning}. Relation networks \cite{sung2018learning} and Prototypical Networks \cite{snell2017prototypical} are examples of such approaches in FSL, enabling models to quickly adapt to new tasks with only a few samples.

Optimization-based meta-learning focuses on 'learning to fine-tune' through good parameter initialization, enabling rapid adaptation to new tasks with minimal gradient updates. MAML prepares a model for fast adaptation to new tasks through a two-step training process, maintaining compatibility with any model that learns via gradient descent \cite{finn2017model}. Reptile simplifies MAML by steering the initialization towards weights that perform well across multiple tasks, effectively rendering it a simpler variant of MAML \cite{nichol2018reptile}. The optimization framework of MAML and its variants provides ideas for the rapid adaptation of EvoFA in the second stage \cite{finn2017model, finn2018probabilistic, nichol2018reptile}.

\section{Methods}\label{methods}

EvoFA achieves rapid calibration of EEG through two-step generalization, mitigating the impact of data modality drift. The main architecture is detailed in Figure \ref{fig2}. FSL-based primary generalization: We employ G2G paired with 2D CNN as the backbone network to extract deep representations from EEG signals and utilize FSL to enhance the model's generalization ability during online calibration. Rapid adaptation secondary generalization: In the meta-testing phase, we cache the data patterns of the source data as they change over time and construct meta-adaptation of the target data relative to the source domain based on time changes, enabling the model to adapt to in new data. The backbone network parameters are denoted by \( \theta \), representing the function \( g_{\theta} \). Extracted features are then processed through a novel adaptation optimization layer \( h_{\phi} \) parameterized by \( \phi \), followed by a classification layer \( c_{W} \) to yield the corresponding emotional output. In the following description, to avoid misunderstandings, we will selectively use the terms source domain data / target domain data and training data / testing data in different contexts.

\subsection{Backbone Network}

G2G offers a novel approach to processing EEG data by capturing the dense information interactions between electrode nodes \cite{jin2023graph}. This allows the processed EEG data to maintain a larger shape and become more suitable for deep learning models, particularly for tasks like emotion recognition where the relationships between electrodes are crucial. Inspired by this concept, we process the EEG feature, denoted as $\mathbf{F} \in \mathbb{R}^{n \times d}$, into a 2D feature interaction matrix $\mathbf{F'} \in \mathbb{R}^{c \times n \times n}$ using the G2G transformation:

\begin{equation}
    \mathbf{F'}^{2D} = \mathrm{G2G}(\mathbf{F}^{1D}),
\end{equation}
where \( n \) represents the number of electrodes, \( d \) denotes the number of features per electrode, and \( c \) indicate channels.

After processing EEG into 2D interaction matrices, we further employ a ConvNet to learn deep representations from these matrices. The ConvNet includes four distinct Conv-BN-ReLU structures. Both the G2G and the subsequent ConvNet are jointly denoted by the representation function \( g_{\theta} \). Next, the data is input into the meta adaptation optimization layer denoted by \( h_{\phi} \). This layer comprises two fully-connected layers and aims to capture the evolving emotional representations during meta-testing in section \ref{metatest}, enabling the model to adapt to domain shift. Ultimately, the classification layer \( c_{W} \) outputs the emotion recognition results. It's worth noting that in the context of RelationNet \cite{sung2018learning}, the classification layer serves a similar function as the relation module.

\subsection{FSL-based Preliminary Generalization}

Traditional deep learning trains models on large datasets, mapping different categories of data to their corresponding labels. FSL, on the other hand, learns distinguishing features from a small amount of data, enabling the model to differentiate between different categories. 

Most EEG emotion recognition methods rely on offline learning with large datasets, rendering them impractical for online use. In contrast, FSL quickly achieves classification ability with a few samples, making it an inherently suitable framework for online emotion recognition.

FSL empowers the model to recognize target data categories in query set $\mathcal{Q}$ using few-shot support set $\mathcal{S}$ calibration samples through episode training. During the training phase, we randomly sample \(N\) classes from the training data. For each class, we randomly select \(K+Q\) labeled samples, where \(K\) are used for the support set and the remaining \(Q\) for the query set (forming an \(N\)-way \(K\)-shot setting). The corresponding support and query sets are:
\begin{equation}
\begin{aligned}
    \mathcal{S}&=\left\{\left(x_{1}, y_{1}\right), \ldots,\left(x_{N \times K}, y_{N \times K}\right)\right\},\\
    \mathcal{Q}&=\left\{\left(x_{N\times K+1}, y_{N\times K+1}\right),\ldots,\left(x_{N\times (K+Q)}, y_{N\times (K+Q)}\right)\right\}.\nonumber
\end{aligned}
\end{equation}

The model is updated by calculating its loss on the query set. In this paper, to demonstrate EvoFA's broad adaptability, we trained three models based on three commonly used FSL frameworks: MatchingNet (MN) \cite{vinyals2016matching}, RelationNet (RN) \cite{sung2018learning}, and ProtoNet (PN) \cite{snell2017prototypical}. The corresponding loss functions are:

\begin{footnotesize}
\begin{equation}
\begin{aligned}
&L_{\text{MN}}(\theta, \phi, W) = -\frac{1}{Q} \sum_{i=1}^{Q
} \sum_{n=1}^{N} y_{in} \log(p_{in}(\theta, \phi, W)) \\
&L_{\text{RN}}(\theta, \phi, W) = \frac{1}{Q} \sum_{i=1}^{Q} (r_{i}(\theta, \phi, W) - y_{i})^2 \\
&L_{\text{PN}}(\theta, \phi, W) = -\frac{1}{Q} \sum_{i=1}^{Q} \log \frac{\exp(-d(f_{\theta, \phi, W}(x_i), n_{y_i}))}{\sum_{n=1}^{N} \exp(-d(f_{\theta, \phi, W}(x_i), n_n))}
\end{aligned}
\end{equation}
\end{footnotesize}

\subsection{From Conventional FSL to EvoFA}

In the basic FSL framework, during the testing phase, the category of unlabeled samples in the query set is determined based on a small labeled support set. The label spaces of the training and testing sets are mutually exclusive, and each episode's support set and query set correspond one-to-one in terms of categories.

Unlike typical FSL tasks, in EEG emotion recognition, due to the limited emotion categories, the labels of the training and test sets are usually identical. Additionally, differences in the distribution of test set data relative to training set data generally stem from modal drift in the subjects' EEG data.

By reducing the modal drift of test data relative to training data, it is expected to improve the model's performance in few-shot testing scenarios. A common approach to reducing modal drift is DA. However, traditional domain adaptation typically requires access to all or a significant portion of the test set data and introduces additional training processes during the training phase. Implementing such domain adaptation contradicts existing online calibration patterns.

EvoFA innovatively introduces test-time adaptation to address the performance degradation caused by modal drift. Before each episode's testing, DA is pre-introduced to adapt the test data to the evolving source data described in Section \ref{metatest}, enhancing the model's adaptability to the test data. EvoFA's introduction offers the following advantages: (1) In the traditional FSL framework, it can be directly introduced during testing, improving the model's recognition ability for test data without additional training. (2) It captures the underlying patterns of modal drift in the data, better addressing unknown test data. (3) It exhibits broad adaptability, enabling its use with various FSL models.

\subsection{Rapid Adaptation Secondary Generalization} \label{metatest}

\begin{algorithm}[tb]
    \caption{Two-step generalisation in EvoFA}
    \label{alg:algorithm}
    \textbf{Require}: The source data \(S\) and the target data \(T\).\\
    \textbf{Parameter}: backbone \( g_{\theta} \), adapter \( h_{\phi} \) and classifier \( c_{W} \). \\
    \textbf{Step 1}: FSL-based Preliminary Generalization
    
    \begin{algorithmic}[1] 
        \FOR{\(t\)=0 \textbf{to} MaxEpoch}
        \STATE update \( g_{\theta} \), \( h_{\phi} \) and \( c_{W} \) with source data \(S\):\\
        \begin{scriptsize}
        \begin{equation}
            \begin{aligned}
            &L_{\text{MatchingNet}}(\theta, \phi, W) = -\frac{1}{Q} \sum_{i=1}^{Q
            } \sum_{n=1}^{N} y_{in} \log(p_{in}(\theta, \phi, W)) \\
            &L_{\text{RelationNet}}(\theta, \phi, W) = \frac{1}{Q} \sum_{i=1}^{Q} (r_{i}(\theta, \phi, W) - y_{i})^2 \\
            &L_{\text{ProtoNet}}(\theta, \phi, W) = -\frac{1}{Q} \sum_{i=1}^{Q} \log \frac{\exp(-d(f_{\theta, \phi, W}(x_i), n_{y_i}))}{\sum_{n=1}^{N} \exp(-d(f_{\theta, \phi, W}(x_i), n_n))}\nonumber
            \end{aligned}
        \end{equation}
        \end{scriptsize}
        \ENDFOR\\
    \end{algorithmic}
    
    \textbf{Step 2}: Rapid Adaptation Secondary Generalization
    
    \begin{algorithmic}[1] 
        \FOR {\(t\)=0 \textbf{to} MaxIter}
            \STATE Sample evolving $S'=\{\mathbf{X}_{s_{1}}, \mathbf{X}_{s_{2}},\ldots,\mathbf{X}_{s_{n}}\}$\\
            \STATE Sample $T' =\{\mathbf{X}_{t_{n+k}}\}$
            \FOR{\(i\)=0 \textbf{to} \(n\)}
                \STATE Compute adapted parameters with gradient descent:\\
                \STATE \begin{scriptsize}\begin{equation}
                    \phi_{i+1} \leftarrow \phi_{i}-\eta_{\text {in }} \nabla_{\phi}\left[d\left(f_{\theta, \phi_{i}, W}\left(\mathbf{X}_{s_{i}}^{spt}\right), f_{\theta, \phi_{i}, W}\left(\mathbf{X}_{t}^{spt}\right)\right)\right]\nonumber
                \end{equation}\end{scriptsize}\\
            \ENDFOR
            \STATE updata \( h_{\phi} \) with:
            \STATE  \begin{scriptsize}
                    \begin{equation}
                        L_{(\phi)}= \frac{1}{n} \sum_{i=1}^{n} d\left(f_{\theta, \phi_{n}, W}\left(\mathbf{X}_{s_{i}}^{qry}\right), f_{\theta, \phi_{n}, W}\left(\mathbf{X}_t^{spt}\right)\right)\nonumber
                    \end{equation}
                    \end{scriptsize}\\
        \ENDFOR
    \end{algorithmic}    
    
\end{algorithm}

Unlike traditional FSL-based methods, which directly measure the similarity between the query set and a small labeled support set to determine the category of samples during testing, EvoFA introduces an additional evolvable fast adaptation module before classification. This module aims to capture potential patterns of modal drift in the source data and apply them for rapid calibration of the target data.

However, while there indeed exists a distribution drift of test data relative to training data caused by modal drift, this change is unknown. To better capture the trend of data changes, we sampled a series of subsets of the source data $S'=\{\mathbf{X}_{s_{1}}, \mathbf{X}_{s_{2}},\ldots,\mathbf{X}_{s_{n}}\}$ on the source domain $S$, and $T'=\{\mathbf{X}_{t_{n+k}}\}$ from the target domain $T$. Each source domain subset represents a snapshot of modal drift to target domain.

MAML learns transferable features by minimizing errors on the support set in the inner loop and errors on the query set in the outer loop. Inspired by this, by gradually narrowing the distance between target data and different subsets of source data in the inner loop, we aim to capture potential patterns of modal drift from source to target. Subsequently, updating the optimizer's parameters in the outer loop is expected to mitigate the damage of modal drift and improve the model's accuracy on the test set.

We constructed different learning patterns for intra and inter-subject tasks, with a more detailed description of both tasks in Section \ref{protocol}. For intra-subject experiments, the modal drift of test data relative to training data gradually intensifies over time. To reduce the impact of modal drift caused by temporal changes, we uniformly sample the training set data over time to form subsets of the source data. For inter-subject experiments, the modal drift of test data relative to training data mainly stems from modal differences between different subjects. To mitigate modal drift caused by different subjects, we sample a subset of data for each subject in the training set.

\textbf{Fast Adapt to Snapshot} In the inner loop, we quantify the drift of each source domain snapshot relative to the target domain and sequentially calculate the MMD loss between the corresponding subset data and the test data. After computing the MMD loss for each subset, we update the parameters of the adaptation layer \( h_{\phi} \). During this stage, the parameters of \( g_{\theta} \) and \( c_{W} \) are fixed. For \(i = 0, 1,\ldots, n-1,\)

\begin{small}
\begin{equation}
    \phi_{i+1} \leftarrow \phi_{i}-\eta_{\text {in }} \nabla_{\phi}\left[d\left(f_{\theta, \phi_{i}, W}\left(\mathbf{X}_{s_{i}}\right), f_{\theta, \phi_{i}, W}\left(\mathbf{X}_{t}\right)\right)\right].
\end{equation}
\end{small}

\textbf{Reducing Drift Re-generalization} To ensure the effectiveness of the representations learned in the inner loop, we aggregate the losses incurred during each sub-domain adaptation and update the \( h_{\phi} \) through gradient backpropagation. This enables the model to learn the trend of changes from the source domain to the target data, thereby enhancing the model's adaptability to the test data.


\begin{equation}
    L_{(\phi)}= \frac{1}{n} \sum_{i=1}^{n} d\left(f_{\theta, \phi_{n}, W}\left(\mathbf{X}_{s_{i}}\right), f_{\theta, \phi_{n}, W}\left(\mathbf{X}_t\right)\right).
\end{equation}

Here, \(d_{\mathcal{H}\Delta\mathcal{H}}\) denotes the ${\mathcal{H}\Delta\mathcal{H}}$ distance.

After obtaining the updated adaptation layer, we perform FSL-based testing.

\section{Experiments and Results}\label{exp}

\subsection{Datasets and Protocols} \label{protocol}

We evaluated the proposed EvoFA on two open-source datasets, SEED \cite{zheng2015investigating} and SEED-V \cite{liu2021comparing}, and conducted both intra-subject and inter-subject experiments on each dataset.

First, we trained the model under three common FSL frameworks and used the standard FSL-based test results as the baseline \cite{vinyals2016matching, sung2018learning, snell2017prototypical}. To demonstrate the effectiveness of EvoFA, we directly used the model trained on the standard FSL framework and introduced the fast adaptation meta-testing module shown in Figure \ref{fig2}(b) during testing.

In this section, all results are derived from the author's replication of the original study, except for the comparison results in Figure \ref{fig3}, which are extracted from the corresponding citations. Due to the differences in tasks and the strict division of the training, validation, and test sets in this study, there are significant discrepancies between the reported results and those in the original papers.

\subsubsection{Datasets}

\textit{SEED dataset}: The SEED dataset collects EEG-based emotion information from 15 subjects, with each subject providing data in three different sessions on separate dates. Each session contains 15 trials, with each trial corresponding to one of three different emotional states: positive, negative, and neutral. In this study, we use preprocessed and feature-extracted EEG signals, each EEG feature corresponds to one second of data and a 3-second sliding window.

\textit{SEED5 dataset}: The SEED-V dataset collects multimodal emotional information from 16 subjects. Each subject provided data in three sessions on different dates, with each session comprising 15 trials. Each trial corresponds to one of five emotional states: happy, disgust, neutral, fear, and sad.

\subsubsection{Protocols} 

\textit{Intra-subject experiment} To simulate intra-subject modal drift, we designed an Intra-subject experiment. For a selected subject, the data from first session was used as the training set, and the data from the subsequent two sessions were respectively used as the validation and test sets. The time intervals between the validation and test sets, relative to the training data, were designed to mimic the effects of modal drift on the subject's EEG data over time.

\textit{Inter-subject experiment} Inter-subject domain drift, a common phenomenon, is extensively present. To simulate this, we designed an Inter-subject experiment. Each subject was sequentially used as the test data, with 12 randomly selected subjects serving as the training data, and the remaining subjects used as validation data. Since each subject's data encompasses three sessions, and there is variability in the data across different sessions for the same subject, we conducted inter-subject emotion recognition experiments separately for each session. The results of these experiments are independently presented in Table \ref{inter-seed} and Table \ref{inter-seed5}.

In both intra-subject and inter-subject experiments, each test was conducted ensuring that the support and query sets were sampled from the same subject's selected session, to maintain a high degree of similarity between the support and query sets.

\subsection{Intra-subject Emotion Recognition}

Intra-subject emotion recognition aims to demonstrate: (1) the advantages of online calibration over supervised learning; (2) the further improvement of EvoFA over standard FSL.

For supervised learning methods, we strictly followed the train-validation-test paradigm. During the training of each model, the model with the best performance on the validation set was saved and tested centrally after all training was completed. The learning rate of the model was set to 0.003 and the batch size was set to 32.

For online calibration methods, we conducted 3-way 1-shot and 3-way 5-shot experiments on the 3-class SEED dataset, where the number of samples per class in the query set was 10. We conducted 5-way 1-shot and 5-way 5-shot experiments on the 5-class SEED-V dataset, where the number of samples per class in the query set was set to five. This was done to ensure that the batch size was close to that of supervised learning.

\subsubsection{FSL-based online calibration}

To demonstrate the superiority of online calibration over supervised learning, we reimplemented four supervised learning-based methods and three FSL-based online calibration methods.

Table \ref{intra-seed} demonstrates that online calibration methods outperform supervised learning methods by over 10\% on the SEED dataset. This substantial difference intuitively reflects the modal drift of a subject's EEG data over time due to data collection from different sessions at different times. Moreover, supervised learning models cannot directly address this modal drift by deepening the network. In contrast, online calibration methods achieve significant improvements over supervised learning models by utilizing a minimal amount of calibration data from the test set.

Table \ref{intra-seed5} reveals an even more substantial performance difference between supervised learning and online calibration methods compared to Table \ref{intra-seed}. Additionally, the model achieves a recognition accuracy of over 86\% in the 5-class task after introducing only one calibration sample for each class. We speculate that this significant performance improvement stems not only from the advantages of the online calibration model but also from the high similarity between data within the same trial in the SEED-V dataset.

\begin{figure}[ht!]
\centering
\subfigure[GCN]{
\includegraphics[width=0.24\textwidth]{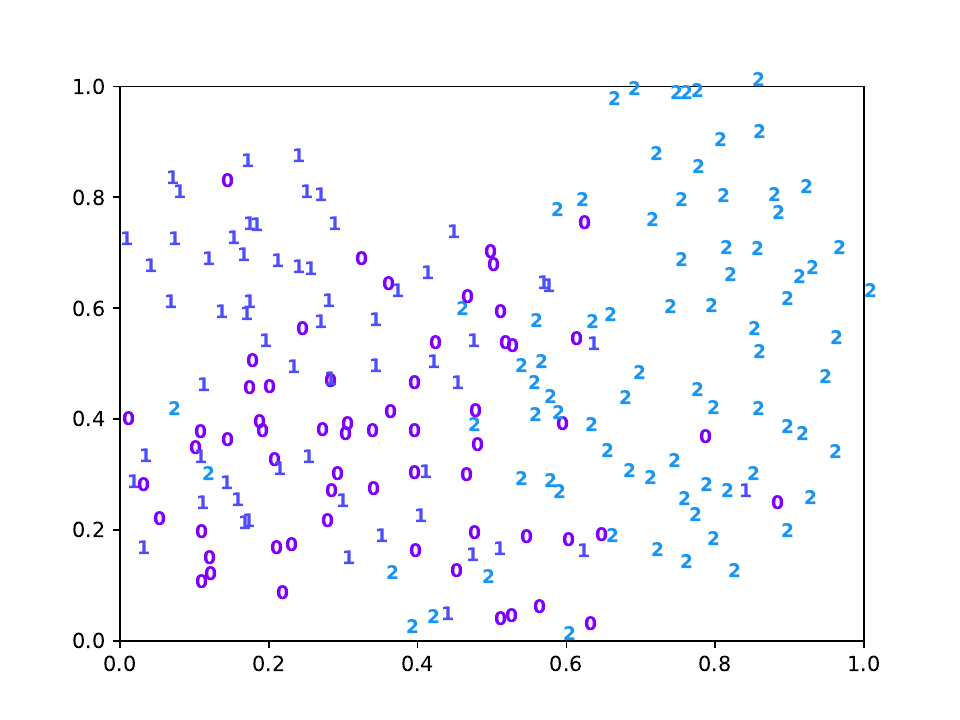}
\includegraphics[width=0.24\textwidth]{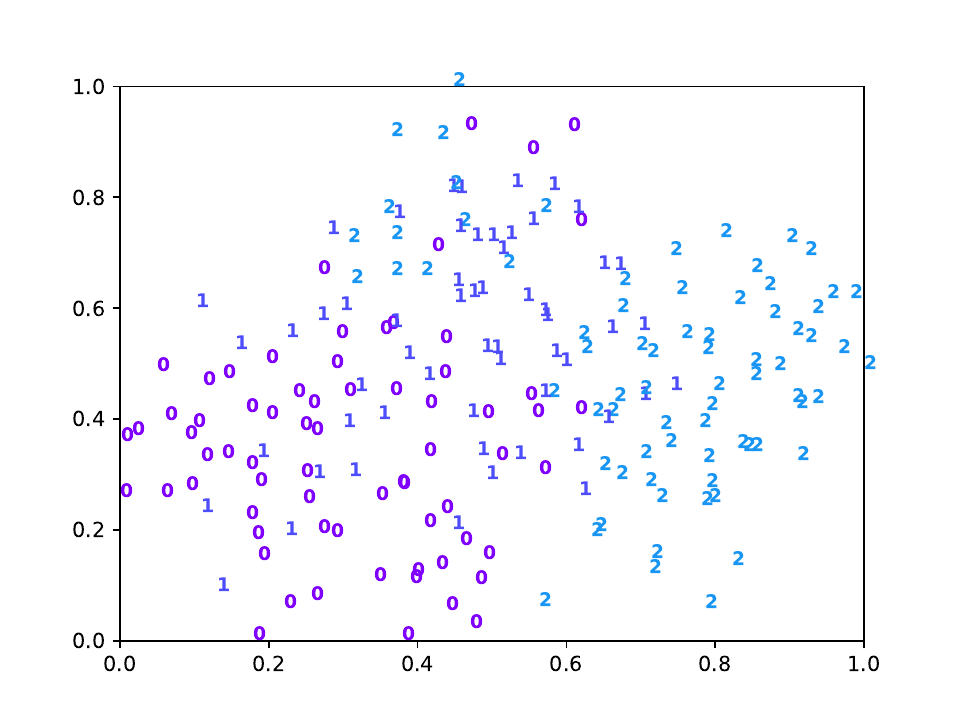}
}
\subfigure[PN + EvoFA]{
\includegraphics[width=0.24\textwidth]{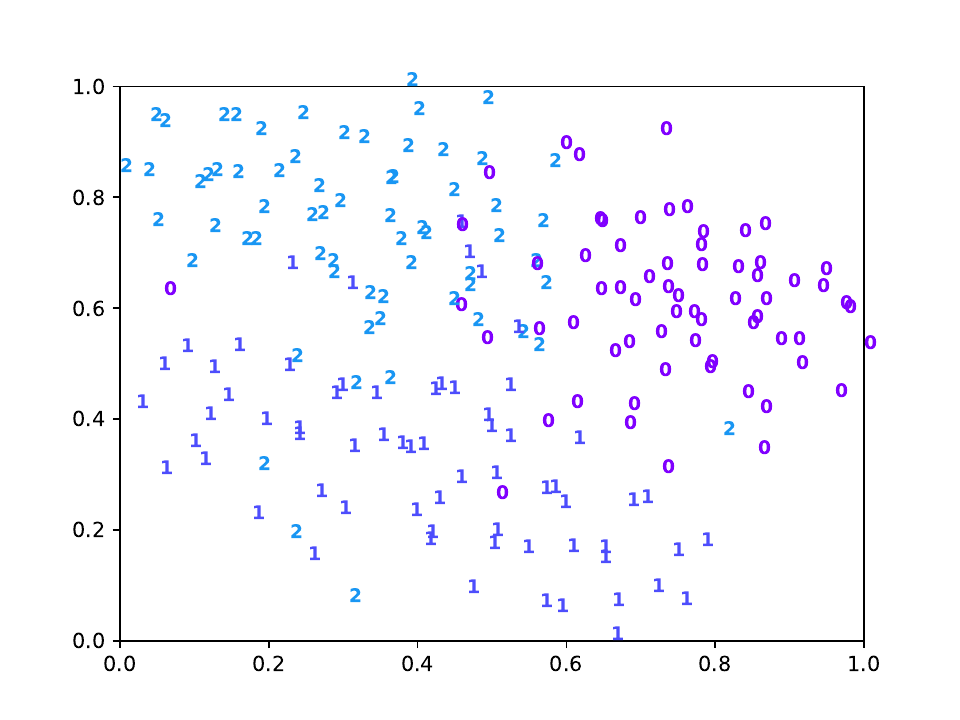}
\includegraphics[width=0.24\textwidth]{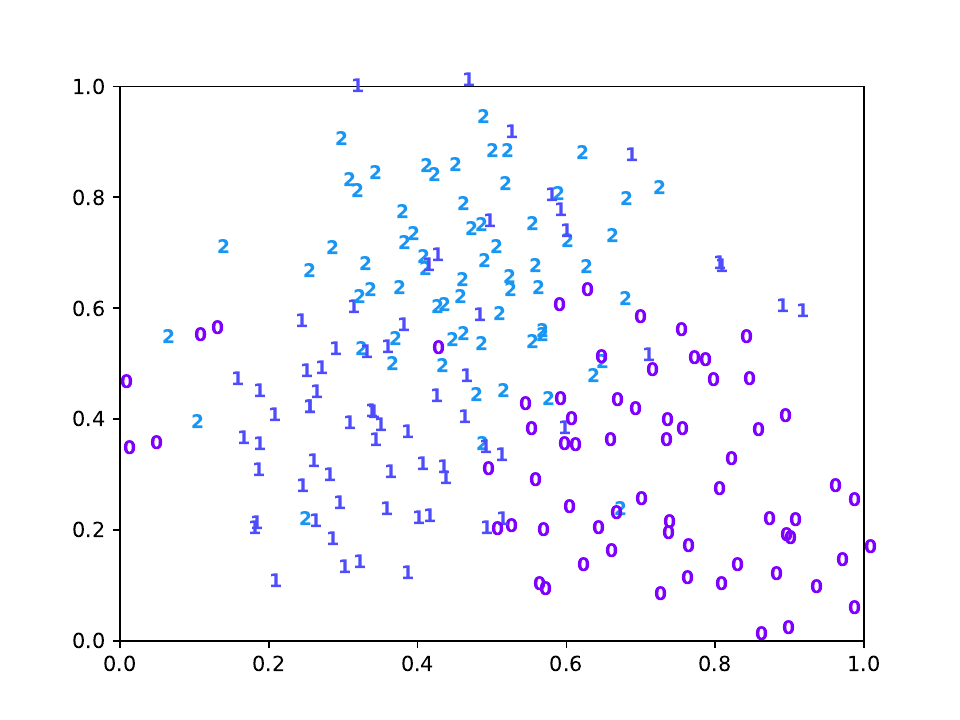}
}
\centering
\caption{t-SNE visualization of emotion recognition output using (a) supervised learning model (GCN) and (b) online calibration model (PN+EvoFA). The different numerical labels correspond to the following emotions: 0-negative, 1-neutral; 2-postive.}
\label{tsne}
\end{figure}

To more intuitively demonstrate the differences in emotion recognition between supervised models and online calibration models, we present the t-SNE visualizations of emotion recognition outputs in Figure \ref{tsne}. These visualizations compare models trained on GCN and PN+EvoFA frameworks for two different subjects. Benefiting from the guidance of calibration data, the online models exhibit better data classification capabilities.

\subsubsection{EvoFA's testing performance boost}

\begin{table}
\centering
  \caption{Intra-subject emotion recognition on the SEED dataset}
  \begin{tabular}{cccc}
    \toprule
     \textbf{Supervised Method} & \multicolumn{2}{c}{\textbf{Acc / Std}}\\
    \midrule
    DBN \cite{hinton2006reducing} &  \multicolumn{2}{c}{ 58.51 / 12.20}\\
    GCN \cite{kipf2016semi} & \multicolumn{2}{c}{ 61.97 / 9.89}\\
    DGCNN \cite{song2018eeg} &  \multicolumn{2}{c}{ 63.85 / 11.04 }\\
    G2G \cite{jin2023graph} &  \multicolumn{2}{c}{ 66.35 / 10.53 }\\
    \toprule
    
    \textbf{Online Method}  &  \textbf{1-shot} & \textbf{5-shot}\\
    \midrule
    MN \cite{vinyals2016matching} & 80.35 {$\pm$ 6.72}   &   84.37 {$\pm$ 5.36}\\
    MN + EvoFA & 81.10 {$\pm$ 6.22}   & 84.92 {$\pm$ 5.12} \\
    \specialrule{0em}{1pt}{2pt}
    RN \cite{sung2018learning} & 79.11 {$\pm$ 10.18}   &   83.97 {$\pm$ 5.35} \\
    RN + EvoFA & 79.26 {$\pm$ 10.17} & 84.89 {$\pm$ 5.66} \\
    \specialrule{0em}{1pt}{2pt}
    PN \cite{snell2017prototypical} & 84.07 {$\pm$ 7.60}   & 87.88 {$\pm$ 6.20}  \\
    PN + EvoFA & 84.27 {$\pm$ 7.82} & 88.15 {$\pm$ 5.91} \\
    \bottomrule
    \end{tabular}
\label{intra-seed}
\end{table}

\begin{table}
\centering
  \caption{Intra-subject emotion recognition on the SEED-V dataset}
  \begin{tabular}{ccc}
    \toprule
     \textbf{Supervised Method} & \multicolumn{2}{c}{\textbf{Acc / Std}}\\
    \midrule
    DBN \cite{hinton2006reducing}  &  \multicolumn{2}{c}{ 46.30 / 14.88}\\
    GCN \cite{kipf2016semi} & \multicolumn{2}{c}{ 48.56 / 10.41}\\
    DGCNN \cite{song2018eeg} &  \multicolumn{2}{c}{ 48.12 / 11.84 }\\
    G2G \cite{jin2023graph} &  \multicolumn{2}{c}{ 48.67 / 14.82 }\\
    \toprule
    
    \textbf{Online Method}  &  \textbf{1-shot} & \textbf{5-shot}\\
    \midrule
    MN \cite{vinyals2016matching} &94.90 {$\pm$ 2.11}&96.88 {$\pm$ 1.73}\\
    MN + EvoFA & 95.10 {$\pm$ 2.63} & 97.72 {$\pm$ 1.52}\\
    \specialrule{0em}{1pt}{2pt}
    RN \cite{sung2018learning} &86.35 {$\pm$ 5.84}&88.02 {$\pm$ 4.04}\\
    RN + EvoFA & 87.23 {$\pm$ 5.48} & 89.13 {$\pm$ 10.18}\\
    \specialrule{0em}{1pt}{2pt}
    PN \cite{snell2017prototypical} & 95.69 {$\pm$ 2.52} & 97.75 {$\pm$ 1.32}\\
    PN + EvoFA & 95.18 {$\pm$ 2.95} & 97.89 {$\pm$ 1.64} \\
    \bottomrule
    \end{tabular}
\label{intra-seed5}
\end{table}

\begin{table*}
\centering
  \caption{Inter-subject emotion recognition on the SEED dataset}
  \begin{tabular}{ccccccc}
    \toprule
    \multirow{2}{*}{\textbf{Method}}  & \multicolumn{2}{c}{\textbf{SESSION 1}}  & \multicolumn{2}{c}{\textbf{SESSION 2}}& \multicolumn{2}{c}{\textbf{SESSION 3}}\\
    \specialrule{0em}{1pt}{2pt}
      & \textbf{1-shot} & \textbf{5-shot}& \textbf{1-shot} & \textbf{5-shot}& \textbf{1-shot} & \textbf{5-shot}\\
    \midrule
    MN \cite{vinyals2016matching} & 60.05 {$\pm$ 6.19}   &   69.42 {$\pm$ 5.92}&56.82 {$\pm$ 7.65}&68.36 {$\pm$ 7.62}&61.11 {$\pm$ 7.79}&71.71 {$\pm$ 6.72}\\
    MN + EvoFA & 60.32 {$\pm$ 6.82} & 69.82 {$\pm$ 6.23} & 56.78 {$\pm$ 8.03} & 68.50 {$\pm$ 7.01} & 61.15 {$\pm$ 8.62} & 72.52 {$\pm$ 5.96} \\
    
    \specialrule{0em}{1pt}{2pt}
    \specialrule{0em}{1pt}{2pt}
    RN \cite{sung2018learning} & 49.21 {$\pm$ 7.03}   &   60.19 {$\pm$ 12.15}&47.36 {$\pm$ 10.35}&59.54 {$\pm$ 10.52}&49.89 {$\pm$ 10.31}&59.22 {$\pm$ 10.94}\\
    RN + EvoFA & 49.87 {$\pm$ 6.61} & 60.40 {$\pm$ 11.37} & 47.08 {$\pm$ 9.82} & 59.45 {$\pm$ 10.73} &50.04 {$\pm$ 10.55}& 59.88 {$\pm$ 10.89}\\ 
    \specialrule{0em}{1pt}{2pt}
    \specialrule{0em}{1pt}{2pt}
    PN \cite{snell2017prototypical} & 57.43 {$\pm$ 7.20} & 73.54 {$\pm$ 7.16} & 55.05 {$\pm$ 10.28} & 71.43 {$\pm$ 9.09} & 58.01 {$\pm$ 11.22} & 76.68 {$\pm$ 10.52}\\
    PN + EvoFA & 57.82 {$\pm$ 7.64} & 74.32 {$\pm$ 6.37} & 55.72 {$\pm$ 9.72} & 72.37 {$\pm$ 8.37} & 58.78 {$\pm$ 10.73} & 77.72 {$\pm$ 9.72} \\
    \bottomrule
    \end{tabular}
\label{inter-seed}
\end{table*}

Observing all the results in Tables \ref{intra-seed} and \ref{intra-seed5}, it is evident that all the online calibration networks performed quite well across every setting on each dataset. We speculate two main reasons for this. First, meta-learning based methods forgo learning to fit the data, instead learning an appropriate metric, which gives them better generalization performance relative to traditional methods. Another reason is that the test data has high similarity within the same trial (corresponding to the same class), which makes it easy for the network to distinguish the category to which the query data belongs.

Among the three FSL frameworks, RelationNet demonstrated a significant disadvantage relative to the other two, with its accuracy on the smaller SEED-V dataset being about 10\% lower than the other networks. We hypothesize that the additional relation module in RelationNet might not be adequately trained on small-sample EEG-based emotion recognition datasets, thus failing to learn an effective metric between the support and query sets. On the SEED dataset, increasing the number of samples per category in the support set from 1-shot to 5-shot led to an approximate 4\% improvement in model performance. Similarly, on the SEED-V dataset, increasing the volume of support set data also resulted in about a 2\% improvement, indicating the positive impact of augmenting calibration data quantity on model effectiveness.

EvoFA enhances the adaptability of the test data by making it conform to the temporal variations of the training data and by narrowing the distance to the overall distribution of the test data during the testing process. This approach results in improved model compatibility with the test data. In the testing phase, the introduction of EvoFA led to performance gains across all three different FSL frameworks. On the SEED dataset, there was an average accuracy increase of 0.47\%; on the SEED-V dataset, the average improvement was 0.44\%. Furthermore, comparing the 1-shot and 5-shot settings, a larger support set contributes more effectively to the optimization of meta-adaptation, thereby resulting in more significant improvements.

Although these enhancements are relatively modest compared to the original FSL frameworks, they are significant given that the model does not require retraining. The incorporation of the rapid adaptation module at the time of use, without introducing additional training losses, is a key advantage. Furthermore, the module's effectiveness across all three FSL frameworks demonstrates its general applicability.

\subsection{Inter-subject Emotion Recognition}

\begin{table*}

\centering
  \caption{Inter-subject emotion recognition on the SEED-V dataset}
  \begin{tabular}{ccccccc}
    \toprule
    \multirow{2}{*}{\textbf{Method}}  & \multicolumn{2}{c}{\textbf{SESSION 1}}  & \multicolumn{2}{c}{\textbf{SESSION 2}}& \multicolumn{2}{c}{\textbf{SESSION 3}}\\
    \specialrule{0em}{1pt}{2pt}
      & \textbf{1-shot} & \textbf{5-shot}& \textbf{1-shot} & \textbf{5-shot}& \textbf{1-shot} & \textbf{5-shot}\\
    \midrule
    MN \cite{vinyals2016matching} & 53.56 {$\pm$ 5.97}   &   61.72 {$\pm$ 5.82}&53.49 {$\pm$ 3.80}
    &62.92 {$\pm$ 4.23}&56.93 {$\pm$ 5.61}&67.37 {$\pm$ 5.93}\\
    MN + EvoFA & 53.72 {$\pm$ 5.88} & 62.26 {$\pm$ 5.63} & 53.01 {$\pm$ 4.23} & 63.32 {$\pm$ 4.82} & 57.04 {$\pm$ 5.98} & 67.52 {$\pm$ 5.83}\\
    \specialrule{0em}{1pt}{2pt}
    \specialrule{0em}{1pt}{2pt}
    RN \cite{sung2018learning} & 49.73 {$\pm$ 8.71} & 58.25 {$\pm$ 8.24} & 49.79 {$\pm$ 8.13} &  58.31 {$\pm$ 8.58} & 47.70 {$\pm$ 7.43} & 59.79 {$\pm$ 7.08}\\
    RN + EvoFA & 49.99 {$\pm$ 8.55} & 58.48 {$\pm$ 8.33} & 49.61 {$\pm$ 7.96} & 58.12 {$\pm$ 9.11} & 48.14 {$\pm$ 7.17} & 60.63 {$\pm$ 7.12}\\
    \specialrule{0em}{1pt}{2pt}
    \specialrule{0em}{1pt}{2pt}
    PN \cite{snell2017prototypical} & 52.87 {$\pm$ 6.87} & 75.37 {$\pm$ 8.37} & 55.62 {$\pm$ 7.43} & 76.60 {$\pm$ 8.87} & 54.72 {$\pm$ 7.11} & 77.49 {$\pm$ 6.37}\\
    PN + EvoFA & 53.17 {$\pm$ 6.63} & 76.16 {$\pm$ 8.58} & 55.38 {$\pm$ 8.20} & 77.26 {$\pm$ 8.42} & 54.82 {$\pm$ 7.42} & 77.82 {$\pm$ 6.18} \\
    \bottomrule
  \end{tabular}
\label{inter-seed5}
\end{table*}

Tables \ref{inter-seed} and \ref{inter-seed5} respectively display the experimental results of inter-subject emotion recognition on the SEED and SEED-V datasets. The volume of data for the support and query sets was set identically to the intra-subject experiments described in the previous section.

The performance of the three FSL frameworks in the inter-subject tasks showed a trend similar to that in the intra-subject tasks, where RelationNet continued to underperform, with its effectiveness significantly lower than the other two frameworks. In 1-shot tasks, MatchingNet exhibited the best performance; however, as the amount of calibration data increased, ProtoNet surpassed the others, achieving significantly better results than the other two frameworks.

Compared to the intra-experiment, the accuracy in the inter-experiment dropped precipitously. Since the query and support sets for testing were both sampled from the same session of a single subject, this eliminates the possibility of significant differences between them. We speculate that the drastic decline in model performance can be attributed primarily to the substantial disparity between the training and testing data.

By incorporating the EvoFA module during testing, the model's test accuracy on the SEED dataset improved by an average of 0.41\%, and on the SEED-V dataset, there was an average increase of 0.26\% in test accuracy. Compared to the intra-subject experiments, the effectiveness of EvoFA in inter-subject tasks decreased, which may also stem from the substantial differences between the training and test data, rendering the model's fine-tuning during meta-testing less effective.

\subsection{Comparison with Existing Online Methods}

\begin{figure}
    \centering
    \begin{overpic}[width=0.46\textwidth]{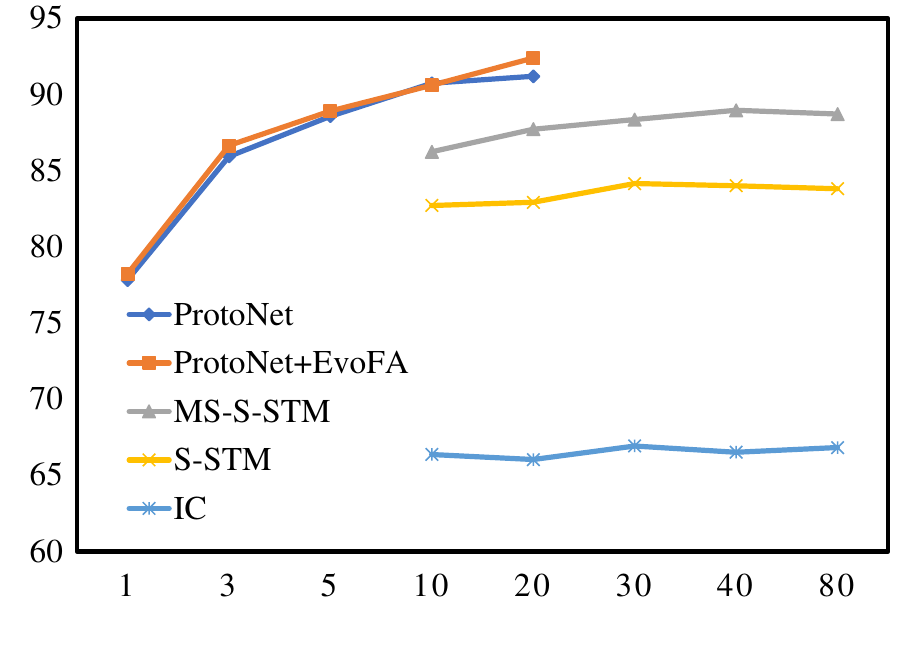}
    \put(-2,29){\rotatebox{90}{\small{{Accuracy (\%)}}}}
    \put(26,0){\small{Calibration data amount (\# - shots)}}
    \end{overpic}
    \caption{Experimental results of five online calibration models with different amounts of calibration data.} 
    \label{fig3}
\end{figure}

To validate the superior advantages of the FSL framework in rapid calibration for EEG-based emotion recognition, we compared ProtoNet with several existing online EEG emotion recognition methods. The results are displayed in Figure \ref{fig3}. MS-S-STM refers to a multi-source online cross-subject emotion recognition method, S-STM represents its version using unified source data, and IC denotes training independent classifiers on calibration data and testing on unlabeled data \cite{li2019multisource}. To ensure fair comparison, we replicated the experimental setup of MS-S-STM. In our experiments, each of the 15 subjects was alternately used as the test set. The first 3 trials of a specified session were used to sample the support set, the following 12 trials for the query set, and the data from the remaining 14 subjects were used as the training set.

It is observable that although the FSL model's test accuracy is only 78\% with just one calibration data per category, increasing the number of calibration data per category to 3 already achieves a recognition accuracy comparable to MS-S-STM using 10 calibration data. When the FSL has five calibration data per category, it surpasses the performance of comparative methods using more calibration data. Moreover, the introduction of the EvoFA module during testing further enhances the model's performance on ProtoNet. These experimental results demonstrate the outstanding advantage of FSL methods over traditional models directly trained with data in the online recognition tasks.

\subsection{Discussion}
Comparing the experimental results of the first four sections, we found that: 

\begin{enumerate}
    \item \textit{Modal shift is widely prevalent.} On the SEED dataset, the performance of the online calibration mode declined by approximately 25\% in the inter-subject tasks compared to the intra-subject tasks; on the SEED-V dataset, this performance drop increased to 40\%. This indicates a more severe data modal difference across inter-subject tasks. In EEG emotion recognition tasks, facing this modal shift is unavoidable.

    \item \textit{Supervised learning encounters bottlenecks.} Faced with such significant modal drift in EEG data, supervised learning has struggled to address this issue through increasing network depth or generalization ability. Although there has been a plethora of work in recent years focusing on improving the accuracy of models on validation sets in EEG emotion recognition tasks, this improvement faces significant shrinkage when additional test sets are introduced.

    \item \textit{FSL still faces limitations.} Current FSL-based online calibration methods still require labeled data. However, it is still costly to calibrate the data category during calibration, so it is very meaningful to update it to an unlabeled online calibration model.

    \item \textit{EvoFA repairs the calibration mode.} EvoFA attempts to address the issue of distribution differences caused by data drift during testing (calibration) by introducing domain adaptation between training and testing data, and preliminary results in experimental results show promise. By introducing testing-time adaptation, models are expected to achieve unlabeled online calibration, which is also our future directions.
\end{enumerate}

\section{Conclusion}\label{conclusion}

This paper addresses the adaptability issue in EEG emotion recognition caused by domain drift and proposes a framework, EvoFA, suitable for rapid calibration in online EEG emotion recognition. To tackle the data modal shift in EEG data and the subsequent need for rapid calibration, EvoFA integrates the rapid adaptation of FSL with the distribution alignment of DA in an organic manner. By adding EvoFA as a plug-in to various FSL frameworks based on online calibration modes, it reduces the domain gap between the source domain data and target domain data during testing, thereby enhancing the model's recognition ability. EvoFA, with its high compatibility and no need for retraining, brings about an additional improvement of around 0.4\% in testing structures across three FSL frameworks. Moving forward, we will continue to focus on rapid modal adaptation in EEG, advancing the practical application of deep models in the EEG domain.

{
\renewcommand*{\bibfont}{\normalfont\footnotesize}
\printbibliography
}

\end{document}